\documentclass[letterpaper, 10 pt, conference]{ieeeconf}  

\IEEEoverridecommandlockouts                              

\usepackage{cite}
\usepackage{graphicx}

\newcommand*{\rom}[1]{\expandafter\@slowromancap\romannumeral #1@}

\title{\LARGE \bf FDM Printing: a Fabrication Method for Fluidic Soft Circuits?}

\author{Savita V. Kendre,$^{1,\dagger}$ Lehong Wang,$^{1,2,\dagger}$ Ethan Wilke,$^{1}$ Nicholas Pacheco,$^{1}$  \\ Loris Fichera,$^{1,3}$ and Markus P. Nemitz$^{1,4,5}$ 
\thanks{This work was supported by the National Science Foundation under CAREER Grant No. 2237506.}
\thanks{Corresponding author: Markus P. Nemitz, mnemitz@wpi.edu}
\thanks{$^{\dagger}$ Indicates equal contribution}
\thanks{$^{1}$Department of Robotics Engineering, Worcester Polytechnic Institute, Worcester, MA 01609, USA.
        {\tt\small }}%
\thanks{$^{2}$Department of Computer Science, Worcester Polytechnic Institute, Worcester, MA 01609, USA.
        {\tt\small}}%
\thanks{$^{3}$Department of Biomedical Engineering, Worcester Polytechnic Institute, Worcester, MA 01609, USA.
        {\tt\small}}%
\thanks{$^{4}$Department of Mechanical and Materials Engineering, Worcester Polytechnic Institute, Worcester, MA 01609, USA.
        {\tt\small}}
\thanks{$^{5}$Department of Electrical and Computer Engineering, Worcester Polytechnic Institute, Worcester, MA 01609, USA.}
}

\begin{document}

\maketitle
\thispagestyle{empty}
\pagestyle{empty}

\begin{abstract}
Existing fluidic soft logic gates for the control of soft robots either rely on extensive manual fabrication processes or expensive printing techniques. In our work, we explore Fused Deposition Modeling for creating fully 3D printed fluidic logic gates.  We print a soft bistable valve from thermoplastic polyurethane using a desktop FDM printer. We introduce a new printing nozzle for extruding tubing. Our fabrication strategy reduces the production time of soft bistable valves from 27 hours with replica molding to 3 hours with a FDM printer. Our rapid and cost-effective fabrication process for fluidic logic gates seeks to democratize fluidic circuitry for the control of soft robots.
\end{abstract}

\section{INTRODUCTION}
Soft robots, fabricated from compliant materials, offer shape adaptability, safer human interactions, and high-impact resilience compared to traditional robots \cite{soft_robot_whiteside}. Despite their pneumatic actuation \cite{pneumatic_acuator}, many soft robots still rely on rigid electronic components like solenoid valves and microcontrollers, compromising system compliance. To mitigate this issue, flexible electronics is increasingly integrated into soft robotic designs \cite{FlexibleElectronics}. Advances in fluidic control have also enabled complex motion patterns \cite{Wehner2016AnRobots, FullPrintTurtle,3DElectroactiveFluidicValve}. Notably, soft bistable valves have emerged as versatile elements for fluidic control, capable of tasks like pressure switching and sensing \cite{Rothemund2018AActuators},\cite{RMGUnderwaterGlider}. These valves have been used in applications like soft crawlers and robotic grippers \cite{Rothemund2018AActuators}, \cite{SoftCompiler}. They can also be configured into various logic gates and circuits, including NOT-gates, AND-gates, OR-gates, SR-latches, D-type latches, 2-bit shift registers, and ring oscillators \cite{Preston2019DigitalDevices}, \cite{Preston2019AOscillator}. With some design adaptations, these devices can serve as memory elements to store 1-bit information \cite{MarkusNemitz2020SoftMemory}. Soft ring oscillators have been used to create a turtle-like, soft-legged robot generating oscillating signals analogous to biological central pattern generators \cite{Tolly_turtle-likerobot}.

\begin{figure} [t]
    \centerline{
    \includegraphics[width=0.49\textwidth]{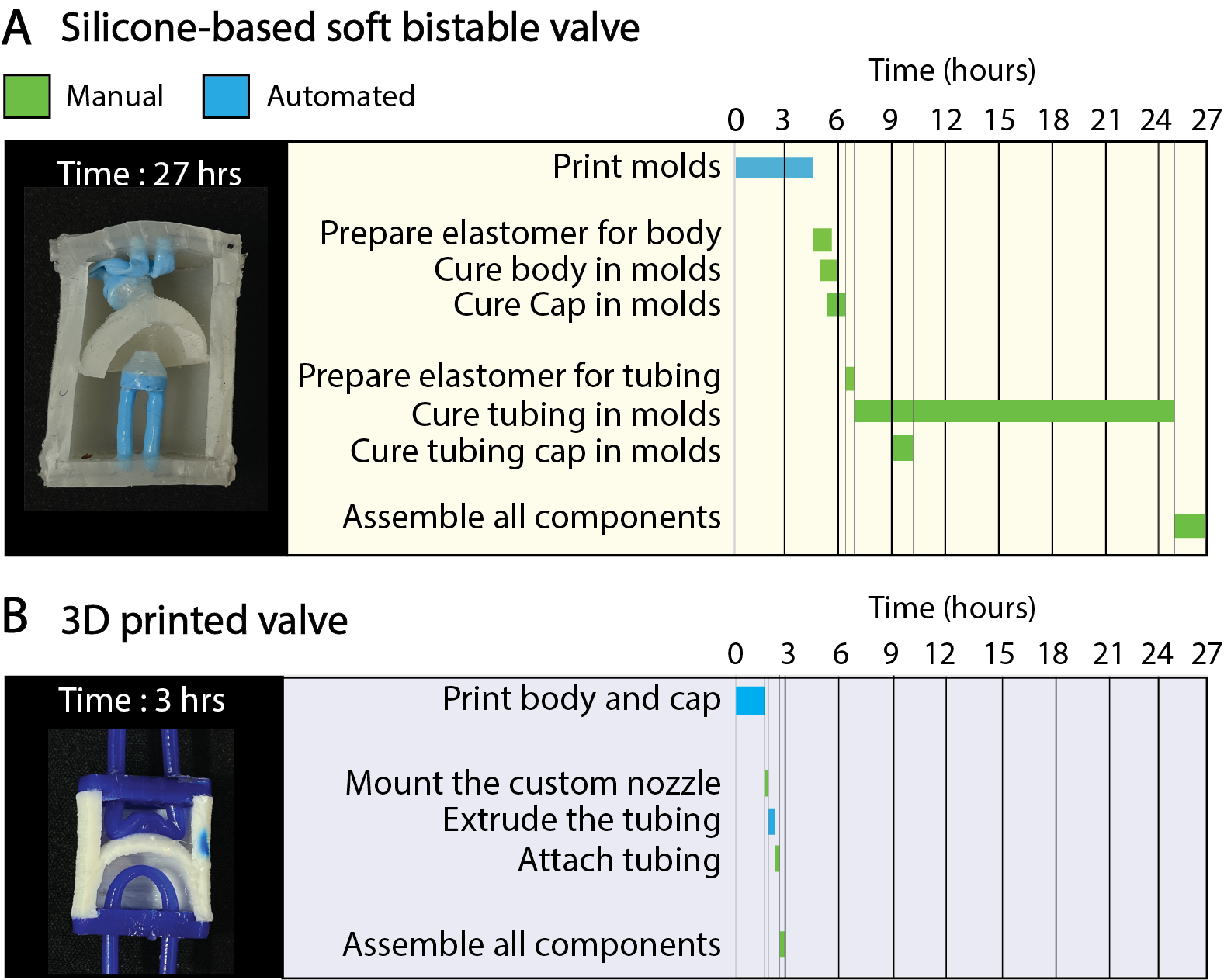}}
    \caption{\textbf{Comparison of replica molding and FDM printing of a soft bistable valve.} The time required for fabricating the soft bistable valve is compared between A) replica molding and B) FDM printing. The fabrication time is reduced from 27 hours to 3 hours when using 3D printing.}
    \label{fig: problem_statement}
\end{figure}

\begin{figure} [t]
\centerline{\includegraphics
[width=0.47\textwidth]{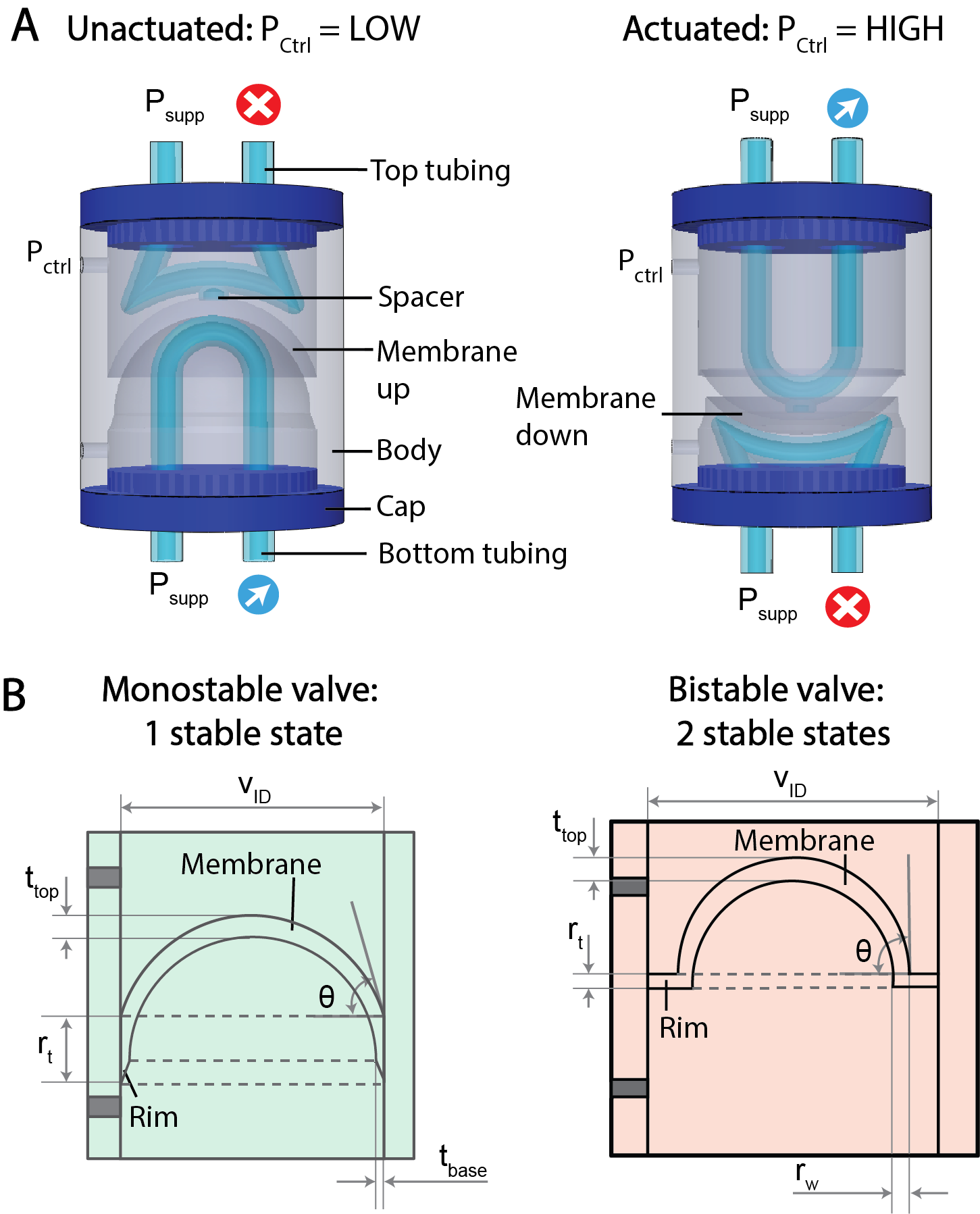}}
\caption{\textbf{The 3D printed valve.} (A) The schematic shows the components of the valve and it's operation in both actuated and unactuated states. (B) The CAD drawing shows the influential parameters and the design of monostable and bistable valves.}
\label{fig: problem_statement}
\end{figure}

The incorporation of soft bistable valves in soft robotic control systems has catalyzed efforts toward economical, automated design architectures. Using commercially available materials, recent work has demonstrated tube balloon logic devices, buckling-sheet inverters, logic-enabled textiles, as well as fluidic computation kits as cost-efficient control modalities \cite{Tracz2022Tube-BalloonElements, decker2022programmable, bucklingsheetoscillator, Logic-enabledtextiles,FluidicComputationKit}. However, these techniques predominantly rely on labor-intensive, manual fabrication processes that are susceptible to operator-induced errors and component variations. As an alternative, additive manufacturing techniques like 3D printing have been employed to develop control elements such as complementary fluidic circuits, tunable soft bistable valves, and fluidic diodes and transistors \cite{JamiePaik2021CMOS-InspiredRobots, TunableValve, FullPrintTurtle}. AirLogic uses a fluid stream deflection mechanism to manipulate airflow, allowing the integration of fluidic logic elements into 3D printed structures \cite{Airlogic}. The introduction of Eulerian path printing now permits the fabrication of airtight actuators and control elements using desktop Fused Deposition Modeling (FDM) technology \cite{Desktopfabricationofmonolithic}. Despite advances toward automated, economical fabrication methods, the required use of high-cost printers and specialized filaments present a financial barrier to widespread replication within the global academic community.

The soft bistable valve serves as a multifunctional element in soft robotic systems. Comprising a cylindrical body segmented by a snapping hemispherical membrane, the valve features two chambers connected by top and bottom tubings to the membrane and end caps. A pressure differential between these chambers triggers the membrane to snap, thereby kinking one tube while un-kinking the other. This operation is functionally analogous to a complementary metal-oxide-semiconductor (CMOS) technology.

The soft bistable valve has been fabricated via soft lithography including replica molding \cite{kendre2022printable}. Silicone rubber or similar elastomers are injected into this mold and allowed to cure. This fabrication process takes up to 27 hours for a single valve (\textbf{Figure 1A}). While cost-effective for mass production, this method is time-intensive for prototyping and iterative designs. Each iteration involves mold redesign, 3D printing, and soft lithography, making it both laborious and time-consuming. Detailed fabrication steps for creating a silicone-based soft bistable valve are available on our GitHub repository (https://github.com/roboticmaterialsgroup/soft-bistable-valve). 

Current fluidic components are hindered by either labor-intensive, manual fabrication or costly, automated machinery. To tackle these issues, we chose the soft bistable valve as a reference and probed cost-effective, automated fabrication via FDM printing—a method noted for its affordability and wide material range. Our research question is: \textbf{Can FDM printing effectively fabricate complementary logic gates using low-cost printers and commercially available thermoplastic polyurethanes?} To this end, we explore the viability of 3D printing a soft bistable valve employing a Prusa MK3S printer (\$649) and Filaflex filament of shore hardness 60A. The key contributions of this work include:
\begin{itemize}
\item The design and fabrication of an entirely FDM-printable soft bistable valve.
\item The reduction of the fabrication time from 27 hours with replica molding to 3 hours using FDM printing.
\item The study and comparison of 3D printed tubes.
\item The introduction of a novel, custom-made 3D printing nozzle for the direct extrusion of tubing.
\item The demonstration of our FDM-printable soft bistable valve as optimized XOR gate and D-latch.
\end{itemize}

\begin{figure} [t]
\centerline{\includegraphics
[width=0.46\textwidth]{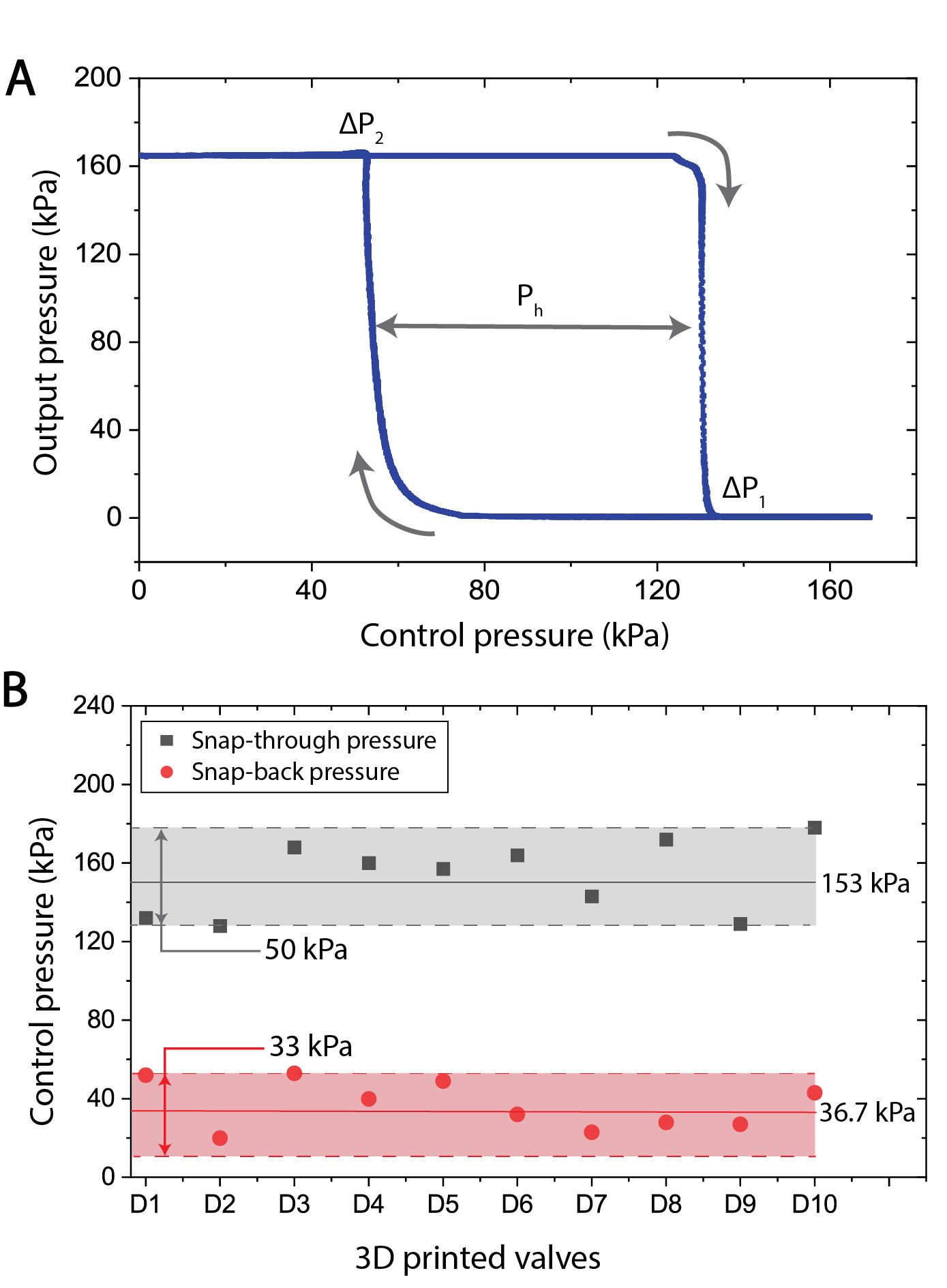}}
\caption{\textbf{Characterization of the 3D printed valve.} (A) The plot shows relationship between the control pressure and the output pressure for the bottom tubing. (B) The variation between snap-through and snap-back pressure for ten 3D printed valves.}
\label{fig: CHARCTERISAITON}
\end{figure}

\section{The 3D printed valve}
Our 3D-printed valve features a body with integrated membrane, top and bottom tubings, and end caps. In its resting state (P\textsubscript{ctrl} = LOW = 0 kPa), the top tubing is kinked, inhibiting airflow, while the bottom tubing remains open, facilitating airflow (\textbf{Figure 2A}). Upon application of positive pressure (P\textsubscript{ctrl} = HIGH = 150 kPa) to the upper chamber, the membrane undergoes a downward snapping motion. This action unkinks the top tubing, allowing airflow, while kinking the bottom tubing, blocking airflow. This dynamic mimics the behavior of a CMOS technology (complementary switching), improving the energy efficiency of the fluidic switch.

\begin{table}[b]
\label{tab:bistab}
\caption{Influence of parameters on the stability of the 3D printed valve (All dimensions are in mm and degree)}
\centerline{\includegraphics
[width=0.48\textwidth]{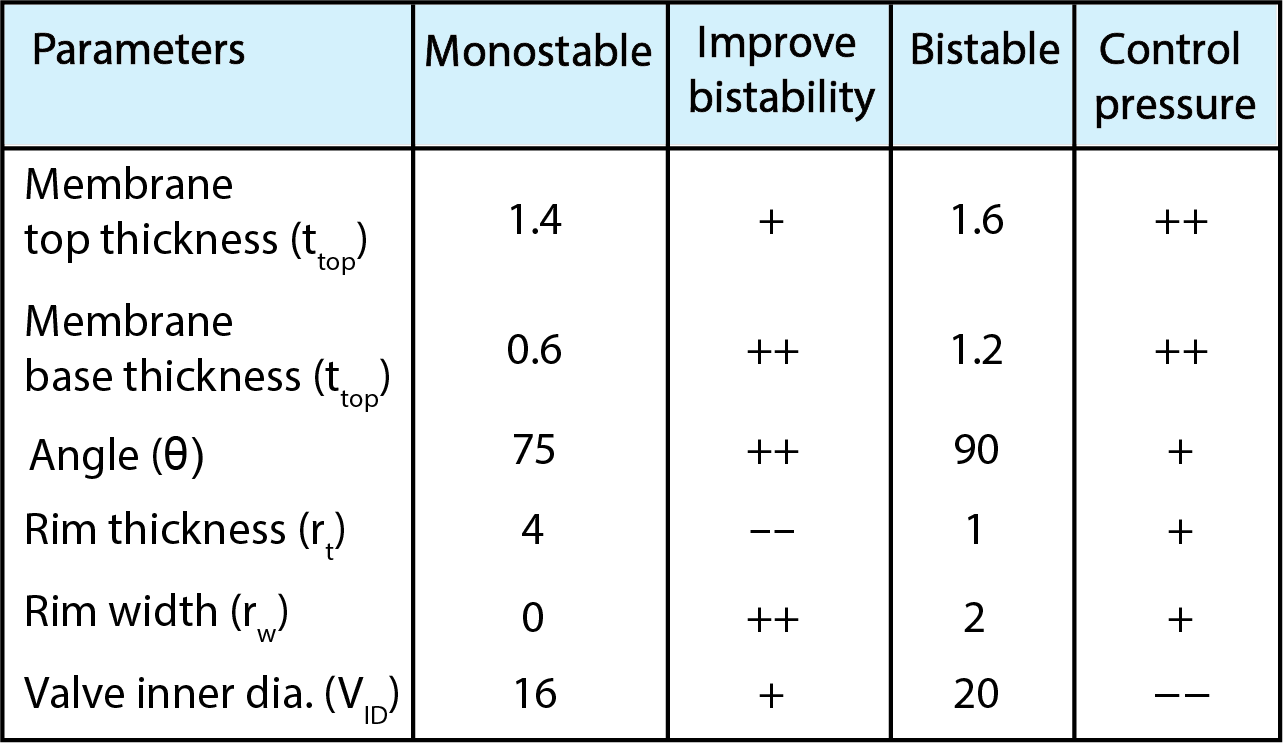}}
\end{table}

We categorized the soft bistable valve from Rothemund et al. into monostable and bistable designs \cite{Rothemund2018AActuators} (\textbf{Figure 2B}). In this context, a monostable valve has a single stable state and requires sustained control pressure for actuation. A bistable valve has two stable states and retains its position even after control pressure removal. Monostable designs are typically used for constructing logic gates, while bistable designs are employed for memory elements. To investigate the parameters affecting bistability, we examined membrane top and base thicknesses (t\textsubscript{top}, t\textsubscript{base}), angle ($\theta$), rim dimensions (r\textsubscript{t}, r\textsubscript{w}), and the inner diameter of the valve (V\textsubscript{ID}). Our data reveal that these factors critically determine the bistability of the valve (\textbf{Table I}). Specifically, we found that increased parameter values generally necessitate higher control pressures for actuation, with the exception of V\textsubscript{ID}.

We studied the relationship between the control pressure and the resulting output pressure (\textbf{Figure 3A}). In our experiments, we applied a constant supply pressure (P\textsubscript{supp} = 160 kPa) to the lower tubing, and atmospheric pressure to the upper tubing. We incrementally increased the control pressure of the valve. The membrane underwent a snap-through at 134 kPa ($\triangle$P\textsubscript{1}), closing the lower tubing. Upon pressure reduction, a snap-back occurred at 56 kPa ($\triangle$P\textsubscript{2}), reopening the lower tubing. The hysteresis of the valve reveals susceptibility to a pressure disturbance (P\textsubscript{h} = 78 kPa) in the system. All plots are based on the monostable design of 3D printed valve.

We evaluated the consistency of snap-through and snap-back pressures across ten bistable valves (\textbf{Figure 3B}). We observe maximum variation of 50 kPa for snap-through pressure and 33 kPa for snap-back pressure; we therefore recommend the actuation of printed valves at 153 kPa. It's crucial to note that this variance could be influenced by multiple factors, including printer type, filament characteristics, and printing conditions. Thus, stringent control of these variables is essential for reproducible, high-quality prints.

To achieve FDM-based fabrication of the valve, we focused on the additive manufacturing of three core components: the body, cap, and tubing.

\begin{figure}[t]
\centerline{\includegraphics
[width=0.5\textwidth]{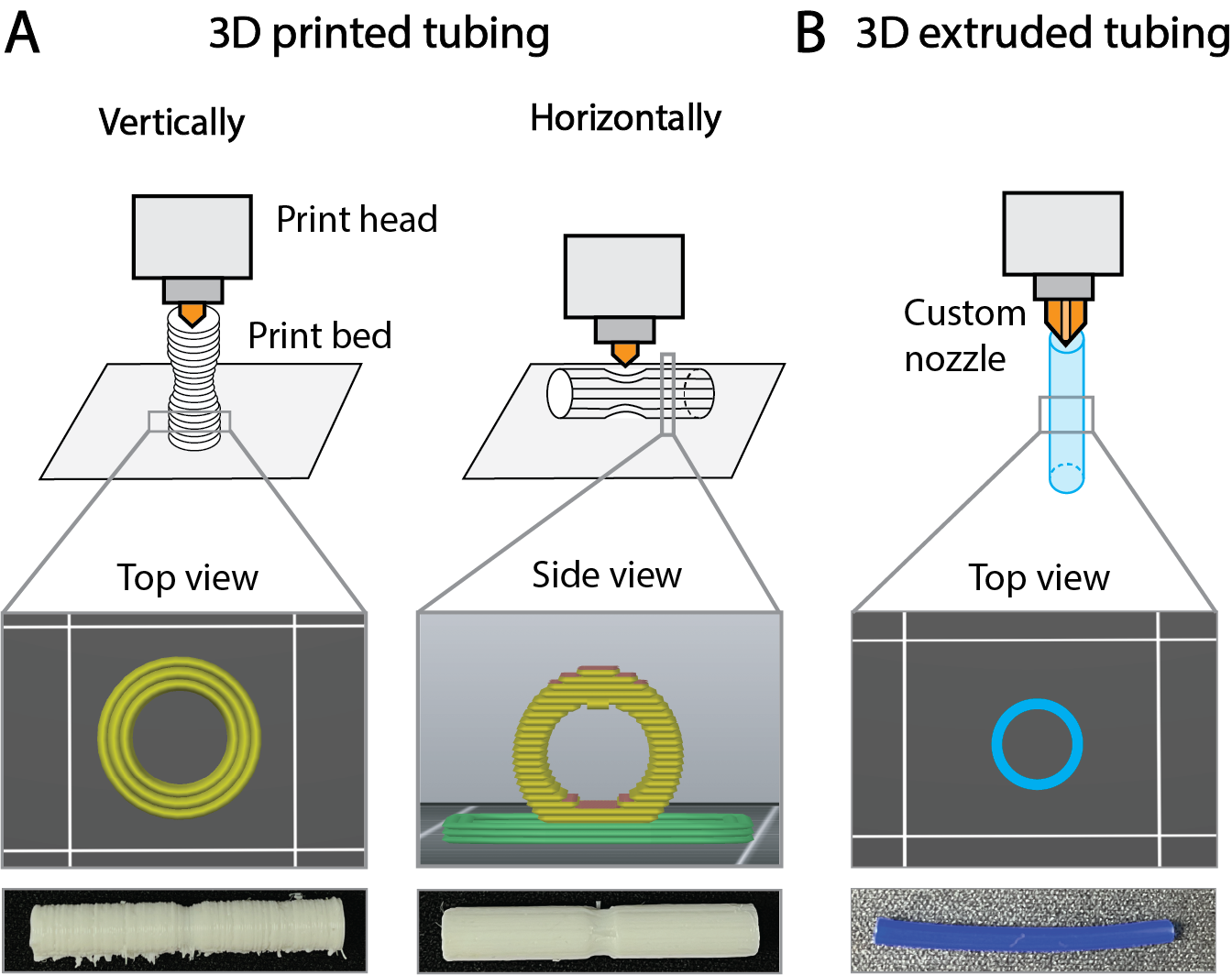}}
\caption{\textbf{FDM printed tubing.} (A) The hourglass-shaped tubing is printed vertically and horizontally. The  vertically printed tubing breaks easily due to weak layer bonding, whereas horizontally printed tubing exhibits the staircase effect leading to minor leakage during kinking. (B) We use our custom nozzle to extrude tubing which has smooth surface finish and increased durability.}
\label{fig: problem_statement}
\end{figure}

\subsection{Fabrication of body and caps}
The 3D-printed valve features a cylindrical body integrated with a membrane. We employed Filaflex 60A and Ninjaflex 85A filaments for the body fabrication. Actuation occurs at 150 kPa with Filaflex 60A, compared to a higher pressure of 220 kPa for Ninjaflex 85A. Therefore, we opted for Filaflex 60A to fabricate the valve body, aiming to reduce the control pressure. Caps can be printed using either filament. The cap design was modified to facilitate bottom-inserted tubing and ensure a press-fit along with adhesive sealant to the valve body. 

\subsection{Fabrication of 3D printed tubing}
We used a desktop FDM printer (Prusa MK3S) with a 0.4 $mm$ nozzle diameter and 0.1 $mm$ layer thickness to fabricate circular tubing in both horizontal and vertical orientations (\textbf{Figure 4A}). An hourglass-shaped design was chosen due to its precise kinking location. Vertically printed tubing attached to the cap showed layer separation due to weak layer bonding, resulting in failure after 10 actuation cycles. In contrast, horizontally printed tubing exhibited a staircase effect due to surface roughness, leading to minor leakage during kinking. We also explored alternative tubing geometries, including circular, rectangular, elliptical, hourglass, triangular, and D-shaped. Hourglass-shaped tubing excelled in kinkability and effectively blocked airflow.

\subsection{Our custom nozzle}

We developed a custom 3D printer nozzle designed for isotropic extrusion of thermoplastic materials in tubular geometries \cite{EthanPatent}, effectively circumventing the layer-stacking limitations inherent in conventional FDM printing (\textbf{Figure 5}). By eliminating inter-layer weak points, the resultant tubes exhibit material properties akin to the raw filament, including enhanced airtightness and structural integrity during bending.

The nozzle comprises an inner cylinder and a nozzle body, the latter mimicking the geometry of a conventional "E3D V6" $1$ $mm$ nozzle but featuring an additional venting hole on the side. The inner cylinder, a press-fit insert, allows for hollow extrusion patterns by directing molten plastic around a needle at the extrusion end. This needle incorporates an air channel to the venting hole, facilitating air ingress into the emerging hollow tube and precluding vacuum formation within the extruded structure.

\begin{figure}[t]
\centerline{\includegraphics
[width=0.49\textwidth]{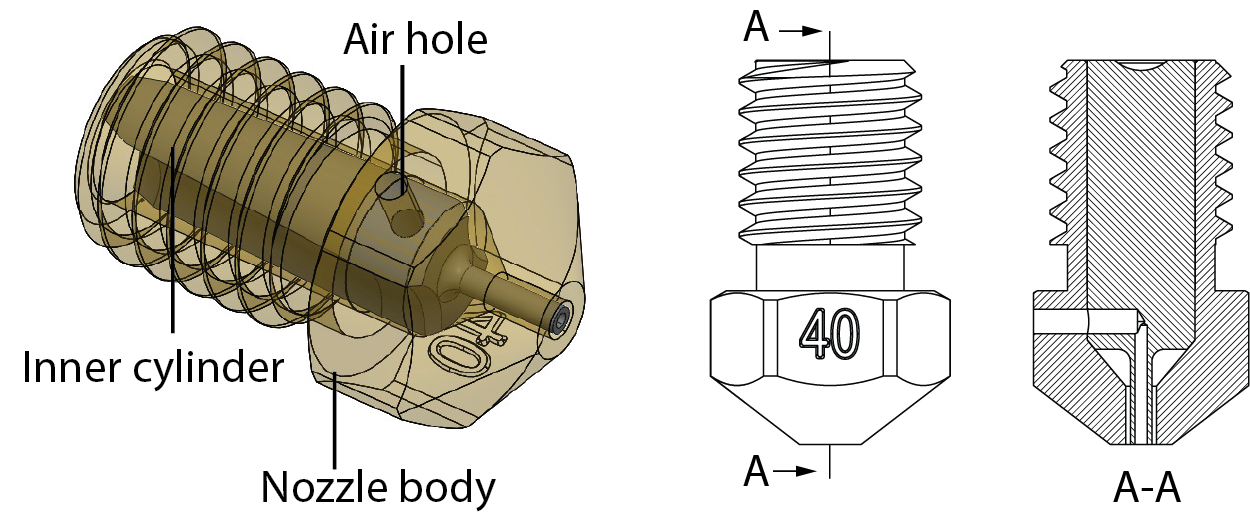}}
\caption{\textbf{ Tube-extruding printing nozzle for FDM printers.} 3D model of a hollow nozzle including a schematic and a cross-sectional view. This nozzle was fabricated in WPI's machine shop; it directly extrudes hollow structures or tubing.}
\label{fig: problem_statement}
\end{figure}

\subsection{Fabrication of 3D extruded tubing}
Our custom nozzle achieves the isotropic extrusion of a tubular structure, which has an inner diameter of 0.7 $mm$ and an outer diameter of 1 $mm$ (\textbf{Figure 4B}).  Due to the standard threading equivalent to an E3D nozzle, our custom nozzle can be used on standard FDM printers. We used a \textit{Prusa Mini+} FDM printer (\$399) and the open-source \textit{Pronterface} software to control the printer. To extrude the tubes, we use G-code commands to configure the print parameters and initiate the tubing extrusion process. Using Ninjaflex 85A, we set the nozzle temperature to 235 $^{\circ}C$ at 100\% fan speed and extruded 1000 $mm$ of filament.

We extruded tubes from multiple materials, settling on Ninjaflex 85A for optimal performance. Post-extrusion, we adhered to our assembly protocol including a unique cutting technique tailored for the kinking mechanism of the valve. Our study covered different material properties and kinking behaviors specific to circular tubes. 

\begin{figure}[t]
\centerline{\includegraphics
[width=0.49\textwidth]{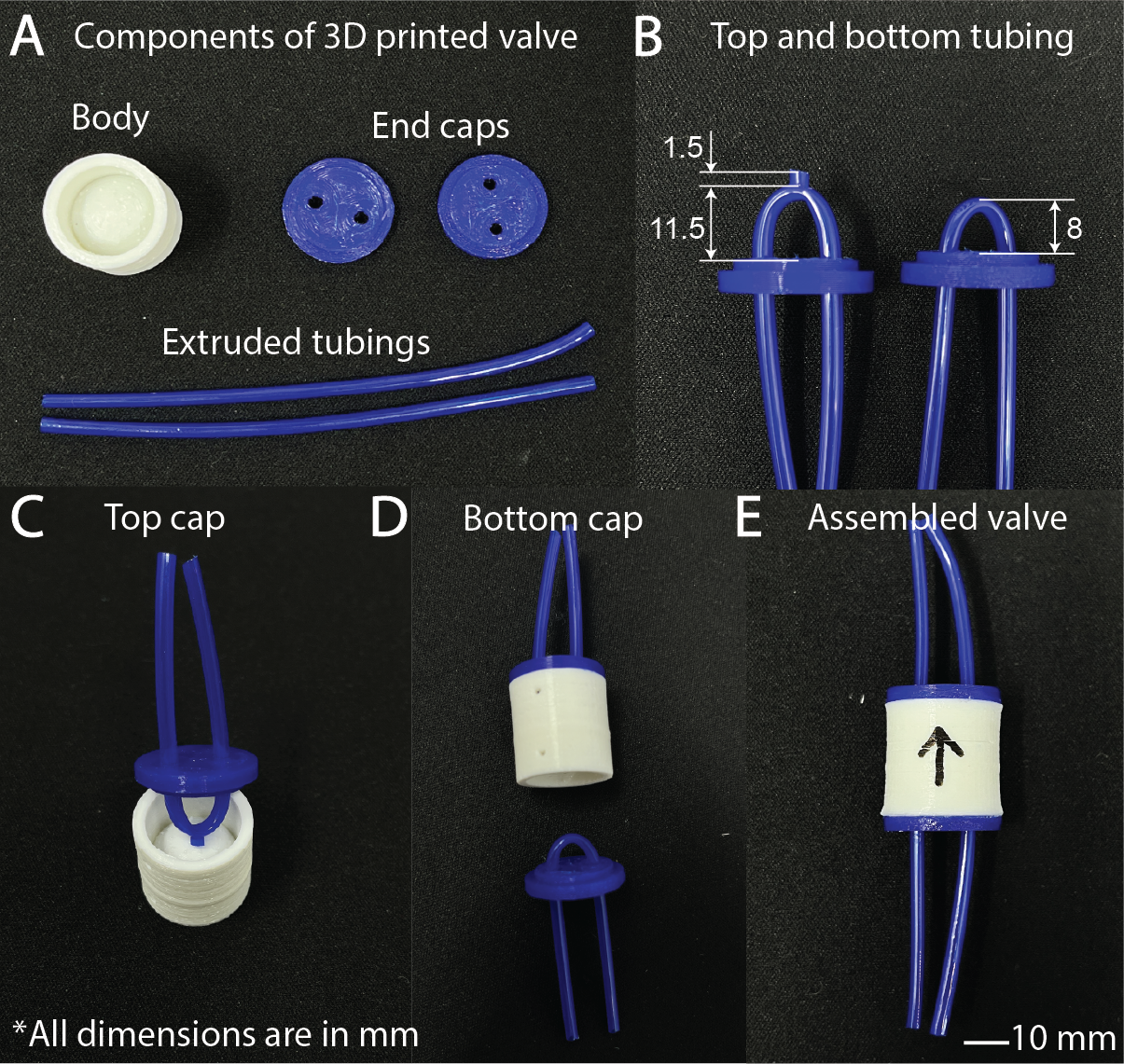}}
\caption{\textbf{The assembly process of the 3D printed valve.} (A) Photographs of all components required to assemble the valve (B) The top tubing of 11.5 $mm$ length along with 1.5 $mm$ spacer and the bottom tubing of 8 $mm$ are inserted through the holes of caps and sealed. (C) The top cap is press-fitted and sealed to the top part of the membrane and (D) the bottom cap is sealed from other end of the body. (E) The fully assembled 3D printed valve.}
\label{fig: assmeblySBV}
\end{figure}

\subsection{Assembly of 3D printed valve}
The assembly process of our 3D printed valve requires a valve body along with two end caps and extruded tubing (\textbf{Figure 6A}). First, we analyze the tubing to verify the isotropic extrusion with constant thickness. For varied thickness, we place the thicker side of the tube inward to minimize the risk of self-induced kinking. We insert the top (11.5 $mm$) and bottom tubing (8 $mm$) through the holes of the end cap, forming an U-shape, and apply adhesive sealant to the connection points (Loctite instant adhesive glue) (\textbf{Figure 6B}). We glue a 1.5 $mm$ piece of tubing (as a spacer) to the top tubing. Once the tubing is attached and sealed to the end caps, we assemble both caps to the body using more adhesive (\textbf{Figure 6C and 6D}). We also apply adhesive to the outer connections to accomplish an airtight system-level seal (\textbf{Figure 6E}). You can find the assembly protocol along with our .stl files of the 3D printable soft bistable valve on our GitHub repository (https://github.com/roboticmaterialsgroup/printed-soft-bistable-valve).

\section{DEMONSTRATIONS}
The soft bistable valve can be configured into different logic gates and circuits as described in previous papers \cite{Preston2019DigitalDevices}, \cite{MarkusNemitz2020SoftMemory}. Similarly, with our 3D printed design, we configure the valve into NOT-, OR-, and AND-gates. These gates are further stacked to implement increasingly complex fluidic circuits.

\begin{figure}[t]
\centerline{\includegraphics
[width=0.5\textwidth]{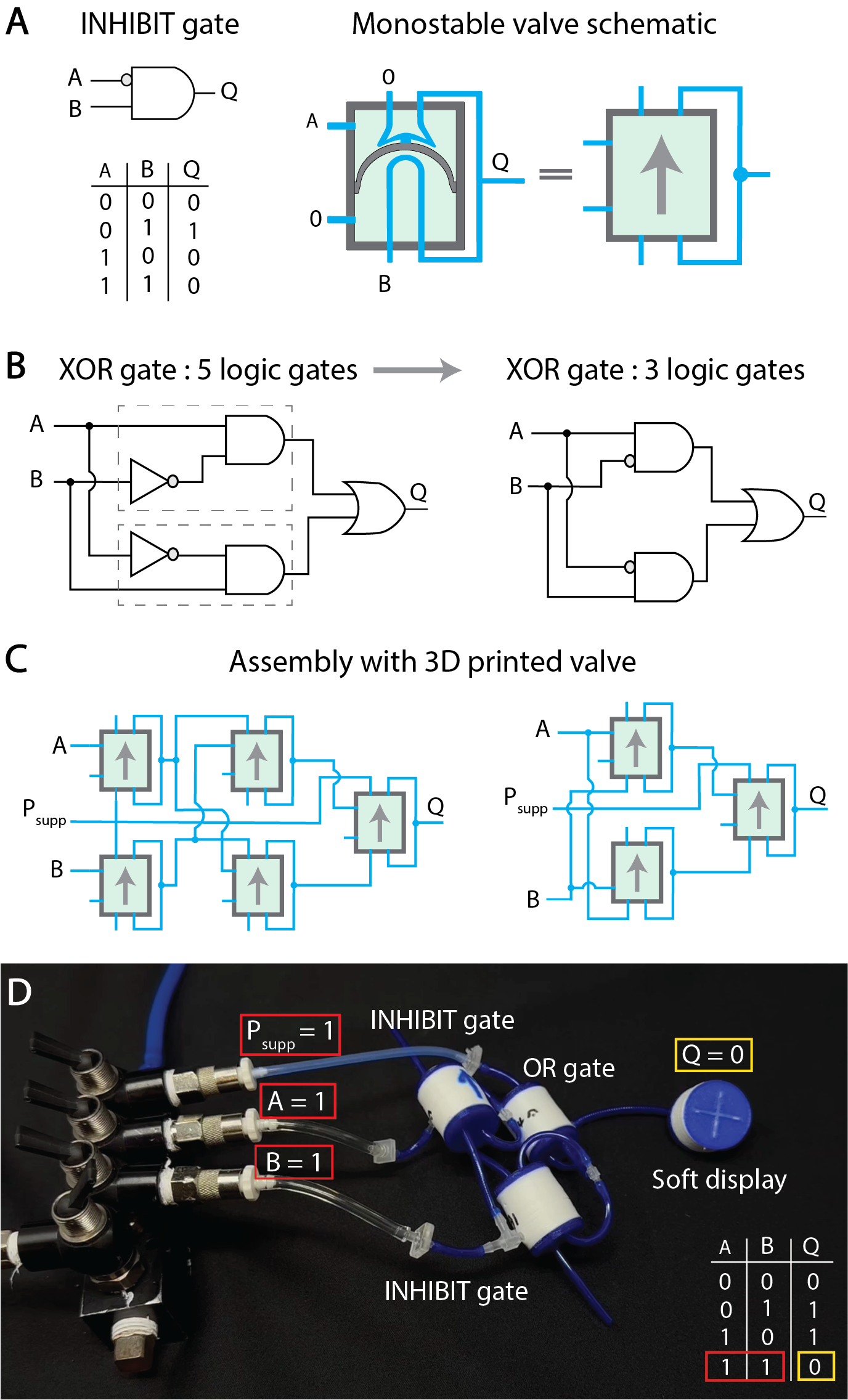}}
\caption{\textbf{Optimised XOR gate.} (A) An INHIBIT gate along with the truth table and schematic representation. (B) The schematic representation of XOR gate with five logic gates. When we optimise the XOR gate using an INHIBIT gate, we only require three logic gates. (C) The assembly of a XOR gate from our 3D printed valves. (D) Implementation of an optimized XOR gate with the output directly connected to a soft display. When both inputs of the XOR gate are HIGH ($A=1$ and $B=1$), the output of the XOR gate is LOW ($Q=0$).}
\label{fig:xor-gate}
\end{figure}

\begin{figure}[ht]
\centerline{\includegraphics
[width=0.5\textwidth]{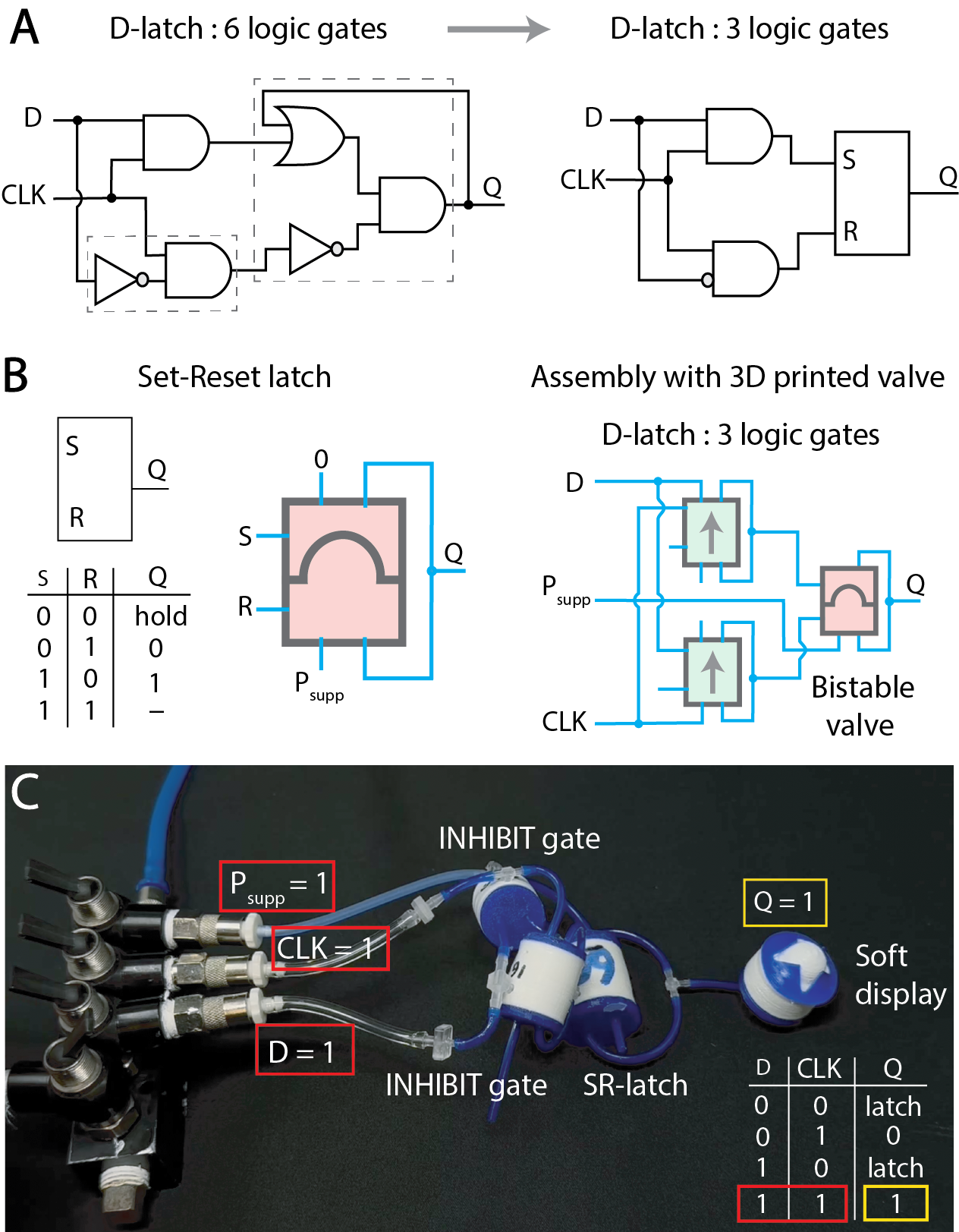}}
\caption{\textbf{Optimized D-latch circuit.} (A) The schematic of a conventional D-latch circuit from six logic gates and the optimized circuit from three logic gates using an INHIBIT gate and a set-reset latch. (B) We modified the 3D printed valve design to create a bistable device; bistable devices act as non-volatile SR latches. (C) Implementation of a D-latch using our 3D printed valves and a soft display. When both the data and clock inputs are HIGH ($D = 1$ and $CLK = 1$), the output turns HIGH ($Q=1$).}

\label{fig:d-latch}
\end{figure}

\subsection{INHIBIT and XOR gates} 
The 3D printed valve can be configured as an INHIBIT gate, which integrates NOT and AND gate operations. By connecting input A to the top chamber and input B to the bottom tubing, the valve outputs \textbf{$Q=1$} only under the condition \textbf{$A=0$} and \textbf{$B=1$} (\textbf{Figure \ref{fig:xor-gate}}). In the schematic of INHIBIT gate, an arrow highlights a monostable valve design along with position of the membrane for comparative analysis. In traditional XOR gate configurations, five logic elements - comprising two NOT gates, two AND gates, and an OR gate - are necessitated. Our work employs an INHIBIT gate to optimize the XOR gate architecture, reducing the required logic gates to three devices only (\textbf{Figure \ref{fig:xor-gate}B and (\textbf{Figure \ref{fig:xor-gate}})C}). Using a printed version of our previously published, silicone-based soft display \cite{MarkusNemitz2020SoftMemory}, we visualized output \textbf{$Q=0$} for inputs \textbf{$A=1$} and \textbf{$B=1$} (\textbf{Figure \ref{fig:xor-gate}}D).
\subsection{Optimized D-latch circuit}
The D-latch, essential in digital systems for data storage and signal synchronization, operates based on two inputs: data (\textbf{$D$}) and clock (\textbf{$CLK$}) signals. The output (\textbf{$Q$}) becomes HIGH (\textbf{$Q=1$}) only when both \textbf{$D$} and \textbf{$CLK$} are HIGH (\textbf{$D=1, CLK=1$}). If the \textbf{$CLK$} signal is LOW (\textbf{$CLK=0$}), the output retains its prior state, irrespective of the D input. When \textbf{$D$} is LOW and \textbf{$CLK$} is HIGH (\textbf{$D=0, CLK=1$}), the output becomes LOW (\textbf{$Q=0$}). This mechanism enables the D-latch to store information.

In traditional D-latch circuits, six logic gates are required: two NOT gates, three AND gates, and an OR gate. Using the INHIBIT gate alongside our 3D-printed bistable valve design, we can optimize the circuit to only three logic gates (\textbf{Figure \ref{fig:d-latch}}). The Set-Reset (SR) latch, a component of the D-latch device, typically uses three logic gates (1 NOT-, 1 AND-, and 1 OR-gate). However, as we demonstrated previously, the silicone-based soft bistable valve can be modified to act as a one-bit, non-volatile memory element \cite{MarkusNemitz2020SoftMemory}. In this work, we replace conventional logic gates with INHIBIT gates and 3D-printed bistable valves (memory elements), reducing the total gate count to three. Our monostable and bistable valve implementations are shown in (\textbf{Figure \ref{fig:d-latch}}). We verified the D-latch function using a soft display, confirming that when both \textbf{$D$} and \textbf{$CLK$} are HIGH (\textbf{$D=1, CLK=1$}), the output is HIGH (\textbf{$Q=1$}) (\textbf{Figure \ref{fig:d-latch}}).

\section{DISCUSSION}

\subsection{3D printing flexible filaments}

FDM printing has limitations, particularly with soft filaments, due to the risk of filament buckling. To mitigate this risk, we employed a Bondtech extruder, enhancing filament feeding and reducing filament slippage. Our elastomeric filaments are highly hydrophilic; we dried them for four hours at 80 $^{\circ}C$ to avoid print defects due to water evaporation at the print nozzle.  We printed all devices at $20$ $mm/s$ speed and a $110\%$ flow rate. While we used a TPU of shore hardness 60A, alternatives include Filaflex 70A, Chinchilla 75A, Ninjaflex Edge 83A, and Ninjaflex 85A. A rise in shore hardness directly correlates with elevated control pressures required to operate the 3D-printable soft bistable valve.

\subsection{Printable Robotics graduate course}

We incorporated these 3D-printed soft bistable valves into a graduate-level course at WPI titled \textit{Printable Robotics}. Equipped with a 3D printer, each student group grasps key concepts related to 3D printing, printable robots, and fluidic control elements. Practical applications are  demonstrated by student groups, such as utilizing the SR-latch to actuate a fully 3D-printed gripper capable of grasping a tennis ball, or configuring a ring oscillator from three 3D printed soft bistable valves to actuate 3D printed, linear actuators sequentially.

\section{CONCLUSIONS}
We explored FDM printing for the fabrication of fluidic valves and circuits towards soft robotic applications. By parallelizing FDM printers, the prototyping and fabrication of fluidic circuits can be accelerated. Our innovative, custom nozzle enables direct 3D tube extrusion with FDM printers. We expect our nozzle to be commercially available in the near future. We demonstrated XOR and D-latch circuits reduced to three logic gates each, employing mono- and bi-stable membranes. Overall, FDM printing emerges as a low-cost approach for the widespread dissemination and adoption of fluidic control elements within the broader academic community.

\addtolength{\textheight}{-8 cm}   






\bibliography{root}
\bibliographystyle{IEEEtran}

\end{document}